\documentclass[spconf]{article}
\usepackage{spconf,amsmath,graphicx,subfigure,amssymb}
\usepackage{booktabs,multirow,multicol}

\def\ie{\textit{i.e.}}

\title{A Robust Ensemble Model For\\Parasitic Egg Detection and Classification}
\name{Yuqi Wang, Zhiqiang He, Shenghui Huang, Huabin Du}
\address{Zhejiang University of Science and Technology\\
Edge Intelligence Security Lab, School of Big Data Science\\
Hangzhou, Zhejiang, China}
\begin{document}
%
\maketitle
\begin{abstract}
Intestinal parasitic infections, as a leading causes of morbidity worldwide, still lacks time-saving, high-sensitivity and user-friendly examination method. The development of deep learning technique reveals its broad application potential in biological image. In this paper, we apply several object detectors such as YOLOv5 and variant cascadeRCNNs to automatically discriminate parasitic eggs in microscope images. Through specially-designed optimization including raw data augmentation, model ensemble, transfer learning and test time augmentation, our model achieves excellent performance on challenge dataset. In addition, our model trained with added noise gains a high robustness against polluted input, which further broaden its applicability in practice.
\end{abstract}
\begin{keywords}
Object detection, biological images, ensemble learning, robustness.
\end{keywords}
\section{Introduction}
\label{sec:intro}
Intestinal parasitic infections (IPIs) remain a main disease severely threatening human health around the world \cite{infect}, leading to a heavy financial burden especially in less developed countries. In WHO's four-part strategy of IPIs controlling, diagnosis acts as a key component and a significant prevention, where stool examination for parasitic ova is one of the most convincing and effective standard in laboratories \cite{WHO02}. However, the requirement of massive examination cost and experienced medical laboratory technologist is still a main obstacles in practice.

With the rise of deep learning technology in computer vision, artificial intelligence becomes an effective solution to many image-related tasks such as image classification \cite{LeNet,ResNet,GoogleNet} and object detection \cite{YOLOv1,YOLOv3,YOLOv5}. The huge success of deep learning technique reveals its broad application prospects in biological fields \cite{bio01,bio02}. Learning from a given set of labeled data, a well trained AI model can easily accomplish certain biological tasks. Google use a deep learning model "in silico labeling (ISL)"\cite{googlemicro} to perform dichroic fluorescent labeling of microscope cell images. Vijayalakshmi and Rajesh \cite{micro1} utilize a transfer learning approach to identify infected falciparum malaria parasite in microscope images.

\begin{figure}[t]
	\centering
	\includegraphics[width=0.85\columnwidth]{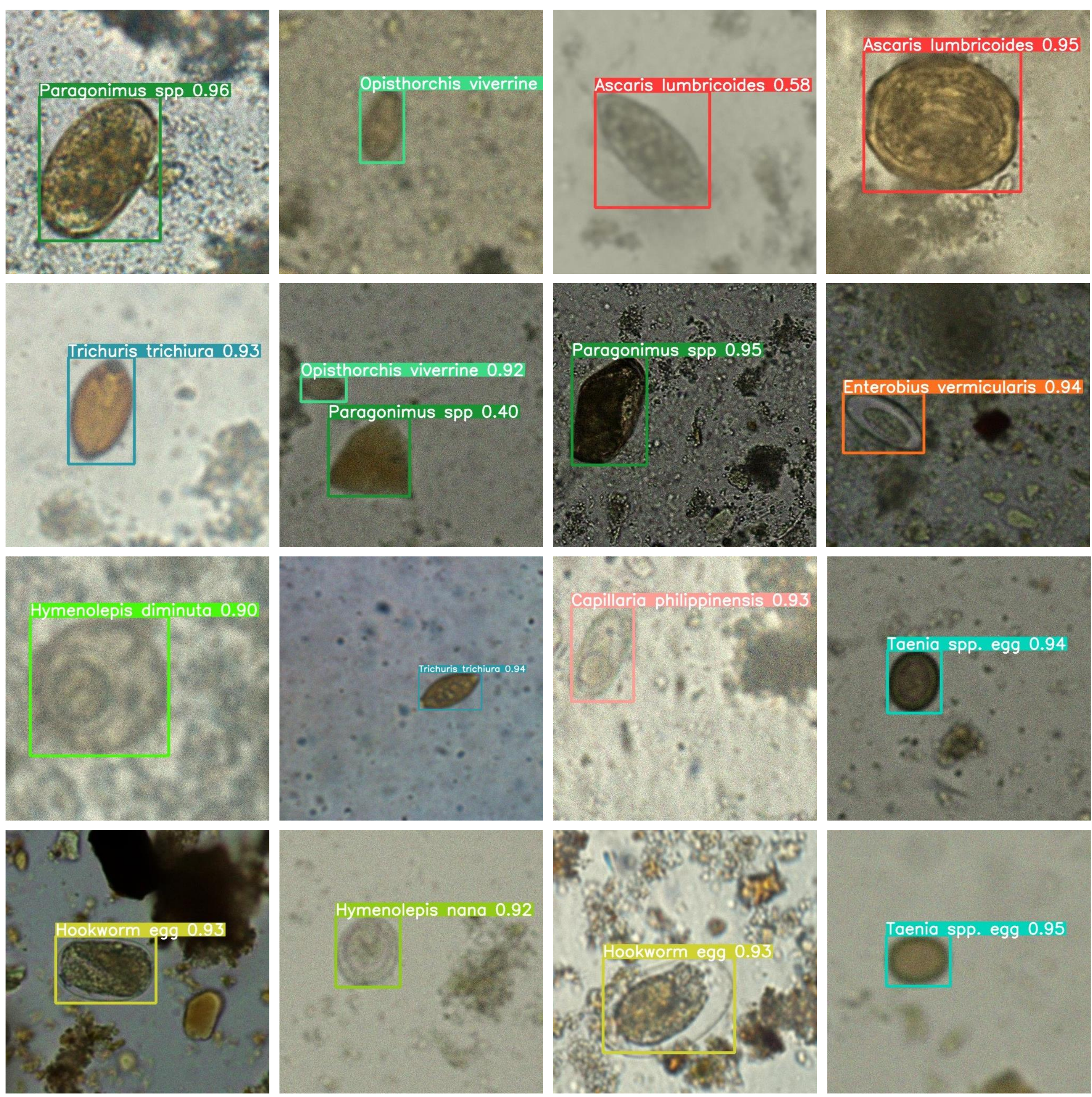}
	\label{fig_intro}
	\caption{Example parasitic egg images detected by our model. Text on detection box represents class of parasite, and number represents prediction confidence.}
\end{figure}

In this paper, we apply YOLOv5 and variant cascadeRCNNs to address the problem of automatic parasitic eggs detection and examination, which are mainstream approaches in object detection tasks. Since these models are not specially designed for biological tasks while differences exist between natural images and biological images \cite{diff}, we make a further improvement of the models' applicability to microscope images, including data augmentation, model ensemble and other training strategies. Considering possible interference to images in practical application scenarios, we also train our model to gain a certain robustness against various noises added on test examples. Experimental results demonstrate that our model achieves excellent performance on test set with more than 0.956 mAP (0.5:0.95). Even with noise-polluted test examples, our model can still achieve 0.967 mAP (0.5).

\section{Basic Algorithm}
\label{sec:frame}
In this paper, we apply YOLOv5 \cite{YOLOv5} and cascadeRCNN \cite{RCNN} as our basic algorithm. The further modification and improvement are accomplished on these models.

\textbf{YOLOv5} is a series of mainstream object detection algorithms developed from \cite{YOLOv1}, among which YOLOv5x has the largest model structure and best performance, making it particularly suitable for detecting tiny objective in high-resolution images.

\textbf{CascadeRCNN} is developed from FasterRCNN \cite{fRCNN}, which presents a cascaded R-CNN structure with different IoU thresholds at different levels. It overcomes the defect that selection of IoU threshold in the R-CNN part has a significant impact on the quality of final detection bboxes. A cascadeRCNN contains three modules: backbone, neck and head with three detection bboxes, where backbone module is replaceable with other feature extractors according to practical requirement.

\begin{figure}[t]		
	\centering
	\subfigure[CLAHE enlarges the contrast between object and background, making features more distinguishable.]{		
		\includegraphics[width=0.8\columnwidth]{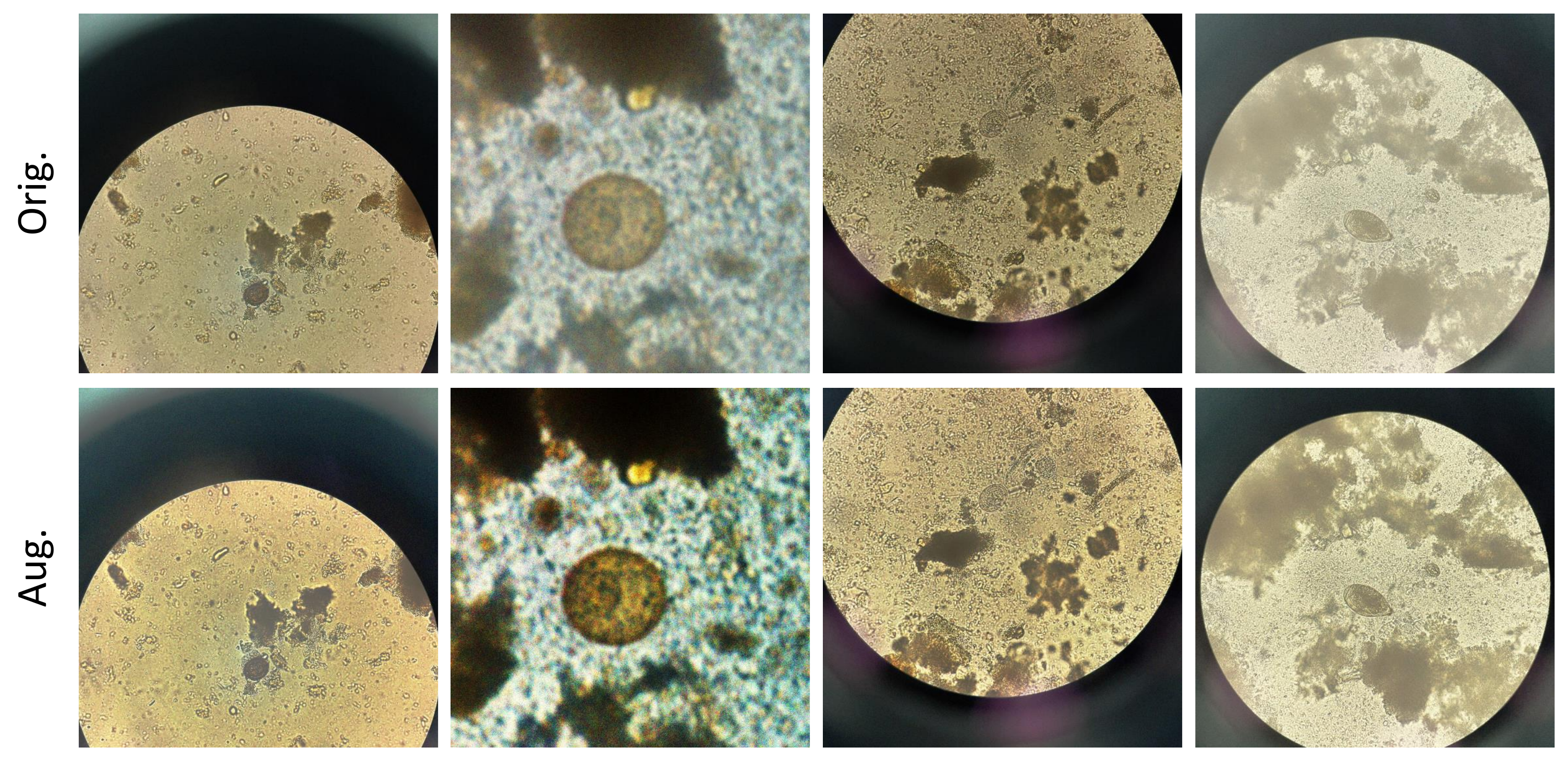}}	
	\subfigure[After CLAHE, the pixel value distribution becomes smoother globally while differentiates partially, which enlarges local features' difference.]{		
		\includegraphics[width=0.95\columnwidth]{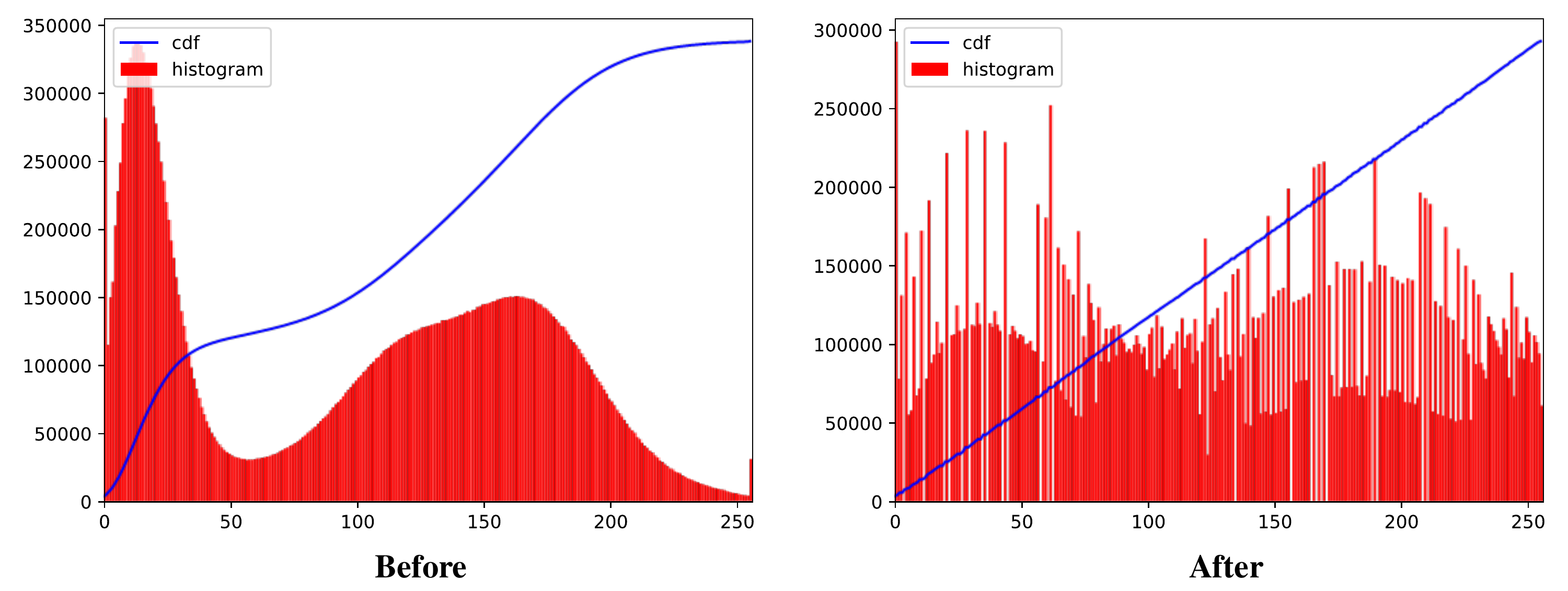}}
	\caption{CLAHE augmentation.}
	\label{fig_aug}
\end{figure}

\section{Data Augmentation}
\label{sec:data}
Since this challenge focus on robustness and accuracy in data-driven technologies, we make various data augmentations in the given dataset. Including normal augmentation, biological data augmentation and robust data augmentation.

\textbf{Normal Data Augmentation.} In standard training procedure, regular data augmentation methods are used to increase models' generalization ability, including resizing, random flipping, normalization, Mixup and padding, etc.

\textbf{Biological Data Augmentation.}
Compared with natural images, microscope images has its unique contribution such as more regular background, more similar objective structure and smaller objective area, which means their features are more indistinguishable between classes. Object detector for natural images may be incapable of effectively accomplishing this challenge. Hence, we augment original data with Contrast Limited Adaptive Histogram Equalization (CLAHE) \cite{CLAHE}, and Enhancement in Mixed Space \cite{mix} to enlarge feature difference between classes, as illustrated in Fig. \ref{fig_aug}. Our augmentation significantly improves model's discriminate ability to object features.

\begin{figure}[t]
	\centering
	\includegraphics[width=0.8\columnwidth]{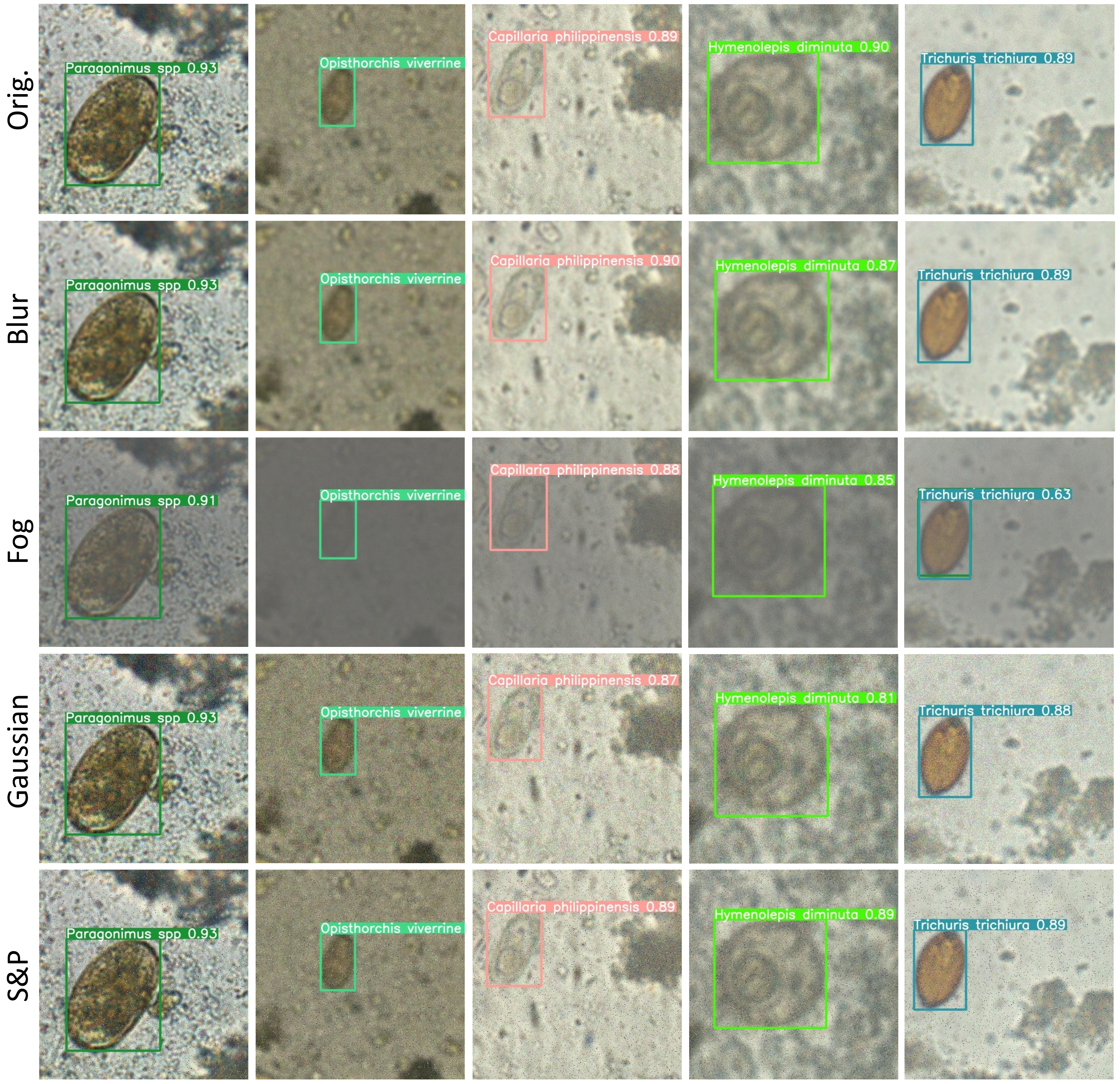}
	\label{fig_noise}
	\caption{Different added noises. Fog and blur can provide a good simulation of practical scenario that microscope lens are out of focus or polluted.}
\end{figure}

\begin{figure*}[t]
	\centering
	\includegraphics[width=0.9\textwidth]{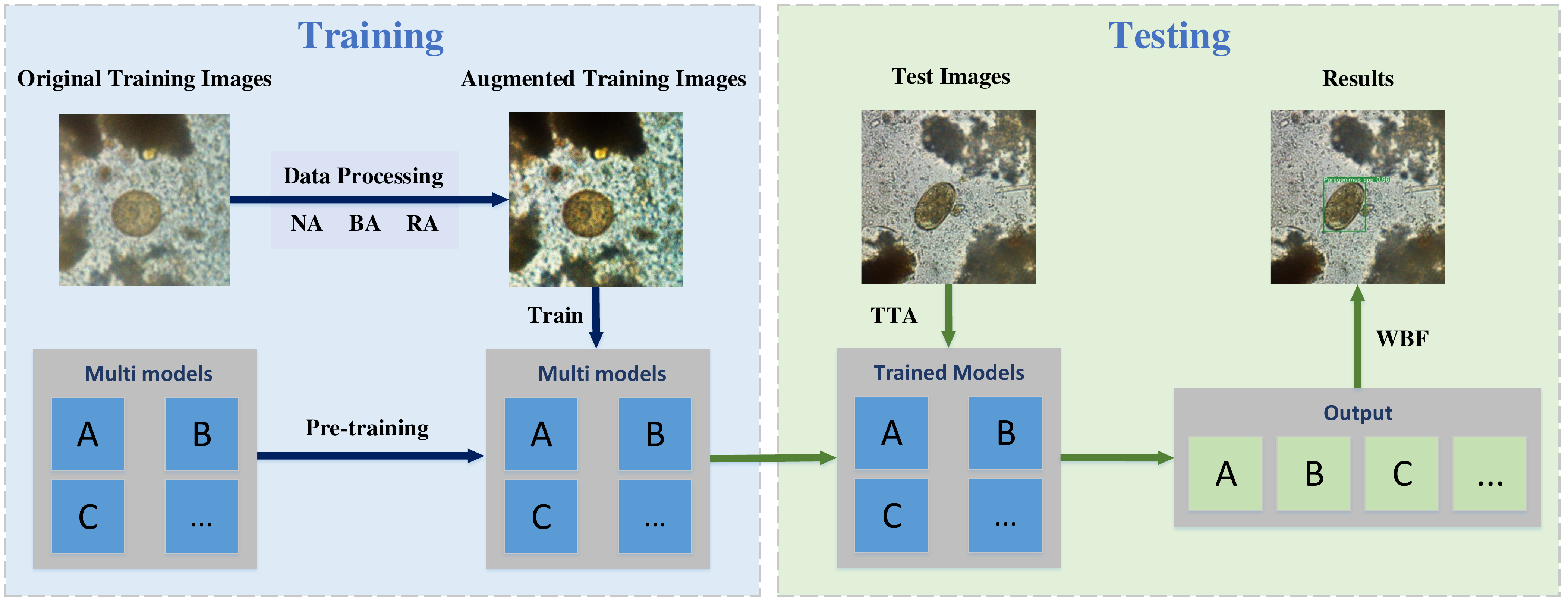}
	\label{fig_procedure}
	\caption{Entire implementation procedure. We train various models on augmented training data and test them on test set with test time augmentation (TTA). Finally, we apply Weighted Boxes Fusion (WBF) to ensemble all models' prediction.}
\end{figure*}

\textbf{Robust Data Augmentation.}
To improve our models' robustness against possible interference occurs on microscope images in practical scenario, our model is fed with both clean and perturbated examples during the training procedure. Multiple perturbation added on images including salt and pepper noise, Gaussian noise, fog and blur. After training with mixed data, our model has gained robustness to various interference, as shown in Fig. \ref{fig_noise}.

\section{Implementation}

\subsection{Pre-processing}

The original dataset contains 11,000 images of 11 classes, each of which contains 1,000 examples. We randomly select 200 examples from each class respectively to construct our validation set with total 2,200 images. The rest of original dataset is our training set, \ie, the training-validation ratio is 4:1. During data processing, we apply the techniques mentioned in Sec. \ref{sec:data} and related parameter settings are listed in Tab. \ref{tab:aug}.

\begin{table}[t]
	\centering
	\caption{Optional augmentation settings.}
	\label{tab:aug}
	\begin{tabular}{cc}
		\toprule
		CLAHE&\texttt{cliplimit}$=8$\\
		\midrule[.5pt]
		Enhancement&\texttt{kernelsize}$=3$\\
		in Mixed Space&\texttt{filtersize}$=5\times5$\\
		\midrule[.5pt]
		Gaussian Noise& \texttt{distribution}$=(0,0.2)$\\
		\midrule[.5pt]
		S\&P Noise& \texttt{number}$=500$\\
		\midrule[.5pt]
		\multirow{3}{*}{Fog}&\texttt{brightness}$=0.4$\\
		~&\texttt{concentration}$=0.03$\\
		~&\texttt{size=imagesize}\\
		\midrule[.5pt]
		Blur&\texttt{kernelsize}$=(6,6)$\\
		\bottomrule
	\end{tabular}
\end{table}

\subsection{Training Procedure}

\textbf{YOLOv5:} We pre-train YOLOv5x and YOLOv5l on Coco dataset \cite{coco}. Then the models are trained on our divided training set. We use SGD with momentum $= 0.937$, and weight decay is $5e - 4$. The initial learning rate is $1e - 2$ and updated by cosine cyclical learning rate strategy. Total training epoch is $300$ with early stop and warm-up epoch is $3$.

\noindent\textbf{CascadeRCNN:} We substitute the backbone in cascadeRCNN with swin-transformer \cite{swin} to better extract object features. The variant cascadeRCNNs are first pretrained on Coco dataset. In training procedure, we use Adam with weight decay $5e-2$. The initial learning rate is $1e-4$ and updated by multi-step decay strategy. Total training epoch is $30$.

\subsection{Post-processing}

When testing the model, we apply a multi-scaled strategy, \ie, the test examples are scaled into several sizes and predicted separately. In addition, we set different confidence threshold for different classes and only retain results whose confidence is larger than its corresponding threshold. If an anchor has a large area in a large-scale image, we consider it as a wrong prediction and discard it. Same principle is applied to small anchor in the small images.

\subsection{Model Ensemble}

After all models are tested, we apply Weighted Boxes Fusion (WBF) \cite{WBF} as our ensemble strategy, which is a state-of-the-art approach of ensemble detection. After ensemble, our model can predict a more accurate anchor on the object with higher confidence. The entire procedure of our implementation is illustrated in Fig. \ref{fig_procedure}.

\begin{table*}[t]
	\begin{center}
		\caption{Experimental results of YOLOv5x on test set. Ticked item represents that corresponding strategy is applied during the training procedure. Aug. represents data augmentation.}
		\label{tab:yolo}
		\begin{tabular}{c c c c c c c| c c c c }
			\toprule
			Normal&Biological&Robust&Pre&Large&Multi&Transfer& \multirow{2}{*}{Precision}& \multirow{2}{*}{Recall}&mAP &mAP\\
			Aug.&Aug.&Aug.&Train&Size&-scale&Learning&&&(0.5)&(0.5:0.95)\\
			\midrule
			&&&&&&&0.931  &0.919  &0.926     &0.860\\
			\checkmark &&&&&&&0.970  &0.962  &0.969     &0.890\\
			\checkmark &\checkmark &&&&&&0.983 &0.973 &0.977 &0.899\\
			\checkmark &\checkmark & \checkmark	&&&&&0.977  &0.961  &0.974 &0.895\\
			\checkmark &\checkmark & \checkmark &\checkmark  &&&&0.989 &0.990 &0.998 &0.921\\
			\checkmark &\checkmark&\checkmark&\checkmark&\checkmark&&&0.992 &0.991  &0.996    &0.927\\
			\checkmark &\checkmark& \checkmark&\checkmark&\checkmark&\checkmark &&0.992  &0.994 &0.998 &0.936\\
			\checkmark &\checkmark& \checkmark&\checkmark&\checkmark&\checkmark &\checkmark  &0.994&0.995 &0.999 &0.945\\
			\bottomrule
		\end{tabular}
	\end{center}
\end{table*}

\begin{table*}[t]
	\begin{center}
		\caption{Experimental results of cascadeRCNN on test set. Ticked item represents that corresponding strategy is applied during the training procedure. Backbone represents substituting the original backbone with swin transformer.}
		\label{tab:cascade}
		\begin{tabular}{c c c c c c c c |  c c c }
			\toprule
			Normal&Biological&Robust&Pre&Back&Large &Multi&Transfer& mAP&mAP  &mAP\\
			Aug.&Aug.&Aug.&Train&bone&Size&-scale&Learning&(0.5)&(0.75)&(0.5:0.95)\\
			\midrule
			&&&&&&&  &0.940  &0.930     &0.858\\
			\checkmark &&&&&&&  &0.963  &0.950     &0.872\\
			\checkmark &\checkmark &&&&&& &0.970 &0.958 &0.880\\
			\checkmark &\checkmark & \checkmark	&&&&&  &0.970  &0.955 &0.881\\
			\checkmark &\checkmark & \checkmark &\checkmark  &&&& &0.984 &0.977 &0.905\\
			\checkmark &\checkmark&\checkmark&\checkmark&\checkmark&&& &0.992  &0.981    &0.931\\
			\checkmark &\checkmark& \checkmark&\checkmark&\checkmark&\checkmark &&  &0.994  &0.987 &0.938\\
			\checkmark &\checkmark& \checkmark&\checkmark&\checkmark&\checkmark &\checkmark  &&0.996 &0.990 &0.945\\
			\checkmark&\checkmark&\checkmark&\checkmark &\checkmark  &\checkmark&\checkmark&\checkmark   &0.999&0.996&0.952\\
			\bottomrule
		\end{tabular}
	\end{center}
\end{table*}

\section{Ablation Studies}
\label{sec:exp}
The main results of two representative models (YOLOv5x and cascadeRCNN with swin tranformer) on test set are summarized in Tab. \ref{tab:yolo} and Tab. \ref{tab:cascade}. We notice that except for robust augmentation, all strategies are beneficial to the model's performance, among which biological augmentation is particularly effective, suggesting that specially designed augmentation methods is necessary in biological-related tasks.

Compared with YOLOv5x, cascadeRCNN's performance is highly related to backbone selection. After substitute its backbone with swin transformer, cascadeRCNN outperforms YOLOv5x. This improvement encourages us to construct an ensemble cascadeRCNN with different backbones for a better performance.

\section{Robustness against Noises}

Though training with perturbated data will slightly damage models' performance on clean test set, it can significantly improves models' performance when test inputs are polluted. We demonstrate our experimental results of different type of perturbations on input images in Tab. \ref{tab:rob}. Compared with YOLOv5 trained without robust augmented data, YOLOv5 trained with augmented data has an improved robustness against a wide range of input perturbation.

\begin{table}[t]
	\begin{center}
		\caption{mAPs (0.5) of YOLOv5 trained with (Y-R) and without (Y-N) augmented data. The first row represents the noise added on test images.}
		\label{tab:rob}
		\setlength{\tabcolsep}{1.5mm}
		\begin{tabular}{c c c c c c }
			\toprule
			Noise &None&Gaussian&Salt \& Pepper& Fog&Blur\\
			\midrule			
			Y-R&0.974&0.972&0.971&0.967&0.969\\
			Y-N&0.977&0.965&0.962&0.944&0.952\\
			\bottomrule
		\end{tabular}
	\end{center}
\end{table}

\section{Final Result}

Final result of our ensemble model on ICIP 2022 challenge of Parasitic Egg Detection and Classification in Microscopic Images is presented in Tab. \ref{tab:final}.

\begin{table}[h]
	\begin{center}
		\caption{Final result. }
		\label{tab:final}
		\setlength{\tabcolsep}{5mm}
		\begin{tabular}{c c c}
			\toprule
			mAP(0.5)&mAP(0.5:0.95)&mIoU\\
			\midrule			
			0.999 & 0.956 &0.966\\
			\bottomrule
		\end{tabular}
	\end{center}
\end{table}
\vspace{-1em}

\section{Conclusion}

In this paper we proposed a practical ensemble model to detect parasitic egg in microscope images. The high performance of our solution to this challenge reveals a high potential of deep learning techniques in the application of biological examination. Through implementation, we conclude that data augmentation is a necessary approach to accomplish biological-related tasks. A series of suitable feature extractors with ensemble learning can effectively improve models' performance. In addition, models' robustness against noises should be another main concern in practical application, since there exists various interference in the real scenario that may ruin the model's inference ability. We expect that our work may provide a brand new baseline for future improvement.


%
%
%


\vfill\pagebreak
\bibliographystyle{IEEE}
\bibliography{Ref.bib}

\end{document}